\documentclass[journal]{IEEEtran}

\usepackage{cite}
\usepackage{graphicx}
\usepackage{url}
\usepackage{wasysym}

\newcommand{\myparagraph}[1]{\vspace{0.1in}\noindent\textbf{#1}}

\begin{document}

\title{Analysis and Observations from the\\First Amazon Picking Challenge}

\author{Nikolaus Correll,~\IEEEmembership{Senior Member},\thanks{N.~Correll is with the Department of Computer Science, University of Colorado at Boulder, Boulder, CO 80309--430.  Phone $+$1\,330\,717--1436, email: ncorrell@colorado.edu.}%
\and{} Kostas E. Bekris,~\IEEEmembership{Member},\thanks{K.~E.~Bekris is with the Computer Science Department of Rutgers University, Piscataway, NJ, USA.}
\and{} Dmitry Berenson,~\IEEEmembership{Member},\thanks{D.~Berenson is with the EECS Department at the University of Michigan, Ann Arbor, MI, USA.}\\
\and{} Oliver Brock,~\IEEEmembership{Senior Member},\thanks{O.~Brock is with the Robotics and Biology Laboratory at the Technische Universit\"at Berlin, Germany.}
\and{} Albert Causo,~\IEEEmembership{Member},\thanks{A.~Causo is with the Robotics Research Centre, School of Mechanical and Aerospace Engineering, Nanyang Technological University, 50 Nanyang Avenue N3-01a-01, Singapore 639798. Phone: $+$65-67905568, Fax: $+$65-67935921, Email: acauso@ntu.edu.sg}
\and{} Kris Hauser,~\IEEEmembership{Member},\thanks{K.~Hauser is with the Department of Electrical and Computer Engineering, Duke University, Durham, NC 27708.}
\and {} Kei Okada, ~\IEEEmembership{Member}, \thanks{K.~Okada is with JSK Robotics Laboratory, the University of Tokyo, Japan.}\\ 
\and{} Alberto Rodriguez, ~\IEEEmembership{Member}, \thanks{A.~Rodriguez is with the Mechanical Engineering Department at MIT, Cambridge, USA.}
\and{} Joseph M.~Romano\thanks{J.~Romano was a member of the advanced research team of Kiva Systems.}
\and{}and Peter R.~Wurman,~\IEEEmembership{Member}\thanks{P.~Wurman was CTO and Technical Co-Founder of Kiva Systems.}
}

\markboth{IEEE Transactions on Automation Science and Engineering (T-ASE), 10.1109/TASE.2016.2600527}{}%


\maketitle
        
\begin{abstract}
This paper presents a overview of the inaugural Amazon Picking Challenge along with a summary of a survey conducted among the 26 participating teams. The challenge goal was to design an autonomous robot to pick items from a warehouse shelf. This task is currently performed by human workers, and there is hope that robots can someday help increase efficiency and throughput while lowering cost. We report on a 28-question survey posed to the teams to learn about each team's background, mechanism design, perception apparatus, planning and control approach. We identify trends in this data, correlate it with each team's success in the competition, and discuss observations and lessons learned based on survey results and the authors' personal experiences during the challenge. 
\end{abstract}

Note to Practitioners:
\begin{abstract}
Perception, motion planning, grasping, and robotic system engineering has reached a level of maturity that makes it possible to explore automating simple warehouse tasks in semi-structured environments that involve high-mix, low-volume picking applications. This survey summarizes lessons learned from the first Amazon Picking Challenge, highlighting mechanism design, perception, and motion planning algorithms, as well as software engineering practices that were most successful in solving a simplified order fulfillment task. While the choice of mechanism mostly affects execution speed, the competition demonstrated the systems challenges of robotics and illustrated the importance of combining reactive control with deliberative planning. 
\end{abstract}


\IEEEpeerreviewmaketitle
\section{Introduction}
The first Amazon Picking Challenge (APC) was held during two days at the
2015 IEEE International Conference on Robotics and Automation (ICRA) in Seattle, Washington. The objective of the competition was to provide a challenge problem to the robotics research community that involved integrating the state of the art in object perception, motion planning, grasp planning, and task planning with the long-term goal of warehouse automation\cite{baker2007exploration,d2012guest}. The competition is in the spirit of a long tradition of competitions as a benchmark for Artificial Intelligence~\cite{anderson2011robotics}. This paper presents the results of a survey of the 26 teams that participated in the challenge and synthesizes lessons learned by the participants and the authors who have either led participating teams or were involved in the challenge organization (Romano and Wurman). 

The diversity of the solutions employed was impressive at a hardware, software and algorithms level. They ranged from large, single robot arms to multiple small robots each assigned to one bin on the shelf, from simple suction cups to anthropomorphic robotic hands, and from fully reactive approaches to fully deliberative sense-plan-act approaches. In surveying the details of each team's approach and questioning them on what they learned from the experience, we hope to extract trends that help us (1)~understand how to eventually solve the problem, and (2)~discover what future robotics research directions are most promising for solving the general problems of perception, manipulation, and planning.

\begin{figure}[!t]
\centering
\includegraphics[width=0.9\columnwidth]{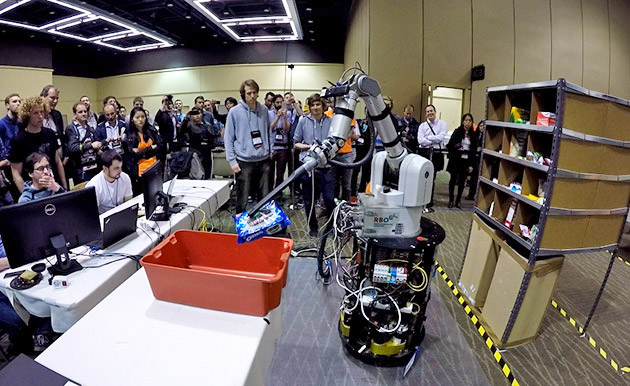}
\caption{The RBO team's robot placing a pack of Oreo cookies that it retrieved from the warehouse shelf into a tote. Image courtesy of RBO team.}
\end{figure}

Extracting such trends, however, is not straightforward. Different teams got comparable results by following almost orthogonal approaches, sometimes stretching the limits of one technology as seen in Table \ref{tab:summary}. Available data on successful grasps, such as removing a specific item from the bin and delivering it to a tote, was sparse. This was in part due to the numerous idiosyncratic ways that complex robotic systems can fail during a single evaluation trial outside of a lab environment. Still, it is possible to make some observations about the strengths and weaknesses of individual approaches, including both mechanisms and algorithms, and how they should be combined to improve the generality of solutions. We can also draw some conclusions about the process. For instance:
\begin{itemize}
\item some of the teams reported that they developed too many components from scratch and did not have time to make them robust, 
\item others reported that the off-the-shelf software components they used as ``black-boxes'' hid important functionality that could not be properly customized.
\end{itemize}
In this regard, there are important lessons about how to simplify the design of complex robotic systems and make them more reliable. 

\subsection{Outline of this paper}

After providing more details on the competition in Section~\ref{sec:rules}, including scoring and rules, Section~\ref{sec:survey} explains the survey and its methodology. The results from the survey, broken into team composition, mechanism design, perception, planning, and summary questions, are described in Section~\ref{sec:results}. Section~\ref{sec:analysis} then contains an analysis of the findings informed in part by the data and in part by the personal experience of the authors. A discussion representing the consensus reached among the authors is presented in Section~\ref{sec:discussion} and Section~\ref{sec:conclusion} summarizes our findings and concludes the paper. 

\section{Overview of the competition and results}\label{sec:rules}

\subsection{The Task}

The APC posed a simplified version of the task that many humans face in warehouses all over the world, namely, picking items from shelves and putting those items into containers. In the case of the APC, the shelves were prototypical pods from Kiva Systems\footnote{Kiva Systems was acquired by Amazon in 2012 and was rebranded Amazon Robotics around the time of this competition.}~\cite{wurman2008coordinating}. The picker was required to be a fully autonomous robot. Each robot had 20~minutes to pick twelve target items from the shelves. 

\begin{figure*}
\centering
\includegraphics[width=0.8\textwidth]{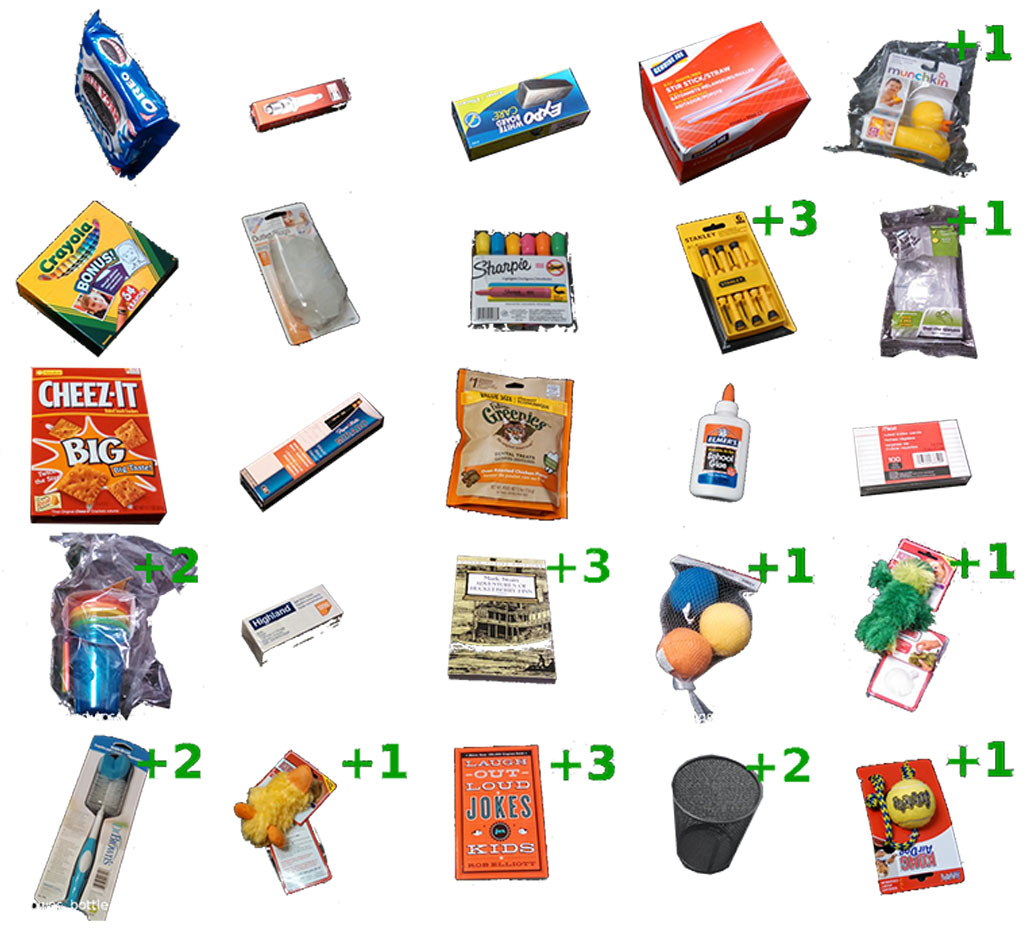}
\caption{\label{fig:apc_items} Items used during the APC competition. Row-by-row starting from the top-left: Oreo cookies, spark-plug, whiteboard eraser, coffee stirrers, rubber ducky, crayola crayons, outlet protectors, sharpies, set of screwdrivers, safety glasses, Cheez-It crackers, set of 12 pencils, cat treats, glue, index cards, set of plastic cups, box of sticky notes, soft cover book, set of foam balls, dog toy, bottle cleaner, dog toy, soft cover book, pencil cup, dog toy. Numbers associated with some items are bonus points awarded for picking difficult items. }
\end{figure*}

\subsection{The Objects}

The items were a preselected set of 25~products, commonly sold on Amazon.com, which would pose varying degrees of difficulty for the contestants' robots. The full set of items is shown in Figure~\ref{fig:apc_items}. Simple cuboids, like a box of coffee stirrers or a whiteboard eraser, were among the easier items to pick. Larger items such as a box of ``Cheez-It'' posed a challenge because it could not be removed from the bin without first tilting it. Smaller items, such as an individual spark plug, were more difficult to detect and properly grasp. The range of cuboid sizes was intentionally chosen to challenge traditional fixed-throw gripper designs that can operate only in a narrow range of possible object widths. 

Beyond size, other items introduced challenges in perception and grasping due to other parameters, such as shape, deformation and the existence of transparent or reflective surfaces. For instance, unpackaged dog toys whose shape varied depending on how items shifted inside their collective packaging, or a pencil cup holder made of black wire mesh that foiled most sensors. Still other items were chosen because they were easy to damage, like two soft-cover books or a package of crushable ``Oreo'' cookies. The books introduced the additional challenge that they could potentially open after lifted from the bin with a vacuum gripper and then collide with the shelf during the retraction process. Objects with reflective covers are challenging for many depth sensors.  

\subsection{The Shelf}

Only the central twelve bins on each pod were used for the contest in order to make the challenge compatible with the reach of the typical commercial armed robot. The organizers of the competition created five stocking arrangements in which the 25~products were distributed among the bins in such a way that each competitor had the same relative difficulty and the same potential to score 190 points. Ten minutes before their trial, each competitor randomly selected one of the five stocking patterns, and the organizers spent the next few minutes arranging the shelf. The team was then given a \emph{.json} (a simple text-based mark-up format) file that contained the names of the items in the bin, and a ``work order'', a simple list containing  the identifiers of the twelve products that needed to be picked. This text file was used as a proxy for data normally provided by a Warehouse Management System (WMS) that is traditionally available in an inventory warehouse setting.

\subsection{Scoring}

The scoring rubric is shown in Table~\ref{table: scoring}. Three of the bins had just the target item, six bins had a target item and one additional distracting item, and the remaining three bins had a target and two (or more) distractions. In addition, some items that were projected to be more difficult to pick were given one to three bonus points each. Points were lost for damaging any item, picking the wrong item (and not putting it back), or dropping the target item anywhere but into the destination tote.

\begin{table}[htb!]
  \begin{center}
    \label{tab:rubric}
    \begin{tabular}{lr}
      Pick target from 1-item bin & 10 pts.\\
      Pick target from 2-item bin & 15 pts.\\
      Pick target from 3+-item bin & 20 pts.\\
      Hard item bonuses & 1--3 pts. \\
      Drop target item & -3 pts. \\
      Each item damaged & -5 pts. \\
      Each non-target item removed & -12 pts. \\
      \multicolumn{2}{r}{} \\
      \end{tabular}
    \caption{The APC scoring rubric with bonuses for more difficult situations and penalties for dropped or wrong items.}
    \label{table: scoring}
  \end{center}
\end{table}

\subsection{Results}

Designers of competitions aspire to create a task that is difficult enough to push the most advanced teams while being accessible to the rest of the field. Based on the results, shown in Table~\ref{tab:scores}, this was achieved in the Picking Challenge. The top team, RBO from the Technische Universit\"at Berlin, picked ten correct items and one incorrect item, for a total score of 148~points. MIT placed second after picking seven items correctly for 88~points. Team Grizzly, a collaboration between Oakland University and Dataspeed Inc., placed third with 35~points and three successful picks. 

\begin{table}[t!]
\begin{center}
\label{tab:scores}
\begin{tabular}{lrccc}
Team & Score & Correct & Wrong & Drops\\ \hline
RBO & 148 & 10 & 1 & 0 \\
MIT & 88 & 7 & 0 & 0 \\
Grizzly & 35 & 3 & 1 & 2 \\
NUS Smart Hand & 32 & 2 & 0 & 0 \\
Z.U.N. & 23 & 1 & 0 & 0 \\
C$^2$M & 21 & 2 & 1 & 0 \\
Rutgers U. Pracsys & 17 & 1 & 0 & 1 \\
Team K & 15 & 4 & 3 & 1 \\
Team Nanyang & 11 & 1 & 0 & 0 \\
Team A.R. & 11 & 1 & 0 & 0 \\
Georgia Tech & 10 & 1 & 0 & 0 \\
Team Duke & 10 & 1 & 0 & 0 \\
KTH/CVAP & 9 & 2 & 1 & 0 \\ \hline
\hline
\multicolumn{5}{c}{} \\
\end{tabular}
\caption{The final APC scores. The 13 teams that scored zero or negative points or did not attempt the competition are not shown in the table.}
\end{center}
\end{table}

Among all 26~teams, a total of 36~correct items were picked, seven~incorrect items were picked, and four items were dropped. About half of the teams scored zero points, including two who set up their robot, but did not get it working well enough to attempt the trials. 
With so few products picked overall, it is perhaps too early to draw meaningful conclusions. But we will offer up some observations. First, the product most commonly picked was the glue bottle, which was successfully picked seven times. This is in part due to the fact that it was alone in the bin in four of five layouts, and paired with only one other product in the fifth. In addition, the bottle was standing upright and was placed inside the bin relatively far from the walls, allowing easy access by grippers. Thus, it had the most favorable arrangements, provided good affordances for picking, and saw the most success. The package of ``Oreos'' and the spark plugs were also targeted in every layout. The cookies were successfully picked only three times (and dropped once) and the spark plug one time (and mis-picked once). The spark plug was in a box, but was still rather small. Surprisingly, the two soft-cover books were picked relatively often (three times each). Several of these picks involved attaching the suction to one cover of the book, which left the pages dangling. A potential complaint is that moving such items using this approach could potentially damage them.

The organizers and participants also ran into some rather mundane, but real-world problems. Despite the fact that 25 items were selected and pre-ordered before the competition, different instances of the same product looked rather different. The rubber duck, for example, sometimes was shipped in a plastic bag and sometimes not. The presence of the plastic bag had a dramatic impact on perception. Similarly, the plastic cups did not always come with the same mix or stacking sequence of colors; sometimes the blue cup was on the outside and sometimes the red cup. Some manufacturers periodically change their product packaging; for example, there were two different sets of artwork for the crayon boxes. When these issues were identified during the competition, the teams were given a choice of which variant to use for their trial, but of course a real industrial system would need to handle these variations automatically.

\section{Survey overview and methodology}\label{sec:survey}

We administered an electronic survey a few weeks after the competition. 
The survey consisted of 28~questions that were grouped into five categories comprising team composition, mechanism design, perception sensors and algorithms, planning and control, and summary questions. We were particularly interested in open source tools that were deemed most important, understanding where most of the development effort was spent, and finding out what respondents thought the biggest challenges were. The survey was administered via \url{surveymonkey.com} and participants were invited by email. The survey questions are also provided in Appendix \ref{sec:fullsurvey}.

We received 31 individual responses from 25 of the 26 teams. 
In the (few) cases teams submitted multiple replies, we checked all quantitative answers for consistency, and manually merged qualitative (text) responses. If quantitative responses were inconsistent, we averaged them. For example, if a team member replied ``Neutral'' to a question, and another ``Strongly agree'', we averaged the team's response to ``Somewhat agree'', and used the value closer to ``Neutral'' for rounding.

\begin{table*}[tb!]
\begin{center}
\label{tab:summary}
\begin{tabular}{l||p{2.5cm}|p{2.5cm}|p{2.5cm}|p{3cm}|p{3cm}}
Team & Platform                                      & Gripper             & Sensor & Perception & Motion Planing \\ \hline\hline
RBO & Single arm (Barrett) ~~ + mobile base (XR4000) & Suction & 3D imaging on Arm, Laser on Base, Pressure sensor, Force-torque sensor  & Multiple features (color, edge, height) for detection and filtering 3D bounding box for grasp selection  & No\\ \hline
MIT & Single arm (ABB 1600ID) & Suction + gripper + spatula & Both 2D and 3D imaging on Head and Arm  & 3D RGB-D object matching  & No\\ \hline
Grizzly & Dual arm (Baxter) ~~ + mobile base (Dataspeed)            & Suction and gripper & 2D imaging at End-effector, 3D imaging for head, and laser for base& 3D bounding box segmentation and 2D feature based localization & Custom motion planning algorithm \\ \hline
NUS Smart Hand & Single arm (Kinova)  & Two-finger gripper & 3D imaging on Robot  & Foreground subtraction and color histogram classification &  Predefined path to reach and online cartesian planning inside the bin using MoveIt. \\ \hline
Z.U.N. & Dual arm (Custom) & Suction & (respondent skipped response) & (respondent skipped response) & MoveIt RRT Planning for reaching motion and use pre-defined motion inside bin \\ \hline
C$^2$M & Single arm (MELFA) on custom gantry & Custom gripper & 3D imaging on End-effector and force sensor on arm & RGB-D to classify object and graspability & No \\ \hline
Rutgers U. Pracsys & Dual arm (Yaskawa Motoman) & Unigripper vacuum gripper \& Robotiq 3-finger hand & 3D imaging on Arm & 3D object pose estimation & Pre-computed PRM paths using PRACSYS software \& grasps using GraspIt\\ \hline
Team K & Dual arm (Baxter) & Suction & 3D imaging on Arm and Torso & Color and BoF for object verification & No \\ \hline
Team Nanyang & Single arm (UR5) & Suction and gripper & 3D imaging on End-effector & Histogram to identify object and 2D features to determine pose & No \\ \hline
Team A.R. & Single arm (UR-10) & Suction & 3D imaging on End-effector & Filtering 3D bounding box and matching to a database & No \\ \hline
Georgia Tech & Single arm & SCHUNK 3 finger hand & 3D imaging on Head and Torso & Histogram data to to recognize and 3D perception to determine pose & Pre-defined grasp using custom software and OpenRave  \\ \hline
Team Duke & Dual arm (Baxter) & Righthand 3 finger hand & 3D imaging on End-effector & 3D model to background subtraction and use color / histogram data. & Klamp't planner to reaching motion\\ \hline
KTH/CVAP & Dual arm + mobile base (PR2) & PR2 2 finger gripper with thinner extension & 3D/2D imaging on head, Tilting laser on Torso and Laser on Base & Matched 3D perception to a stored model & Move to 6 pre-defined working pose and use MoveIt to approach and grasp object \\ \hline
PickNick & Single arm (Kinova) on custom gantry for vertical motion & Kinova 3 finger hand & fixed pair of 3D imaging sensors & 3D bounding box-based segmentation & MoveIt! RRT for motion generation and custom grasp generator\\ \hline
SFIT & Multiple miniature mobile robots on gantry & Custom gripper & 2D imaging and distance sensor & 2D features and color & Visual servoing\\
\hline
\hline
\multicolumn{5}{c}{} \\
\end{tabular}
\caption{Summary of the strategy taken by selected teams}
\end{center}
\end{table*}

\section{Survey results}\label{sec:results}

\subsection{Team composition and background}

The 25~teams that participated in the survey comprised a total of 157~people, that is around 6-7~per team on average. Of these, 79 were graduate students (50\%), 30~undergraduates (19\%), 23~professional engineers (15\%), and 25~in the ``other'' category (16\%), which included post-docs and advising faculty. Most teams exhibited a mix of these groups, with a heavy focus on graduates and post-graduates. The winning team (RBO, academic, Technische Universit\"at Berlin) consisted of seven graduate students and one undergraduate, the second-place team (MIT, academic, Massachusetts Institute of Technology) consisted of five graduate students and a professional engineer, and the third-best team (Grizzly, academic/commercial, University of Oakland/Dataspeed Inc.) consisted of three professional engineers, one graduate student, and two undergraduates. Of the 25 teams, 21 were exclusively affiliated with an academic institution and one identified as a private party (Applied Robotics). Three teams were affiliated with both an academic and a commercial entity (C$^2$M, Grizzly, Robological). 

\begin{figure}
\includegraphics[width=\columnwidth]{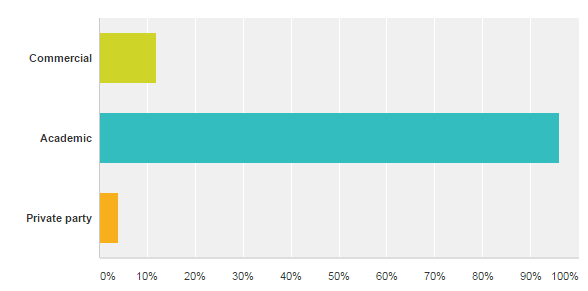}
\caption{Academic, non-academic and private party team composition.\label{fig:teamcomposition}}
\end{figure}

Some teams consisted of a collaboration between a robotics and a vision research group, or in case of the MIT team, a perception company. Asked about missing skills within the team, most teams identified one or more specific skill sets. We analyzed their text replies, and identified the following clusters, ordered by number of occurrences in parentheses: ``computer vision'' (9), ``mechanical design'' (4), ``motion planning'' (4), ``grasping'' (4), ``force control'' (3), ``software engineering'' (2), and ``visual servoing'' (2). 

\subsection{Platform}

\begin{figure}
\includegraphics[width=\columnwidth]{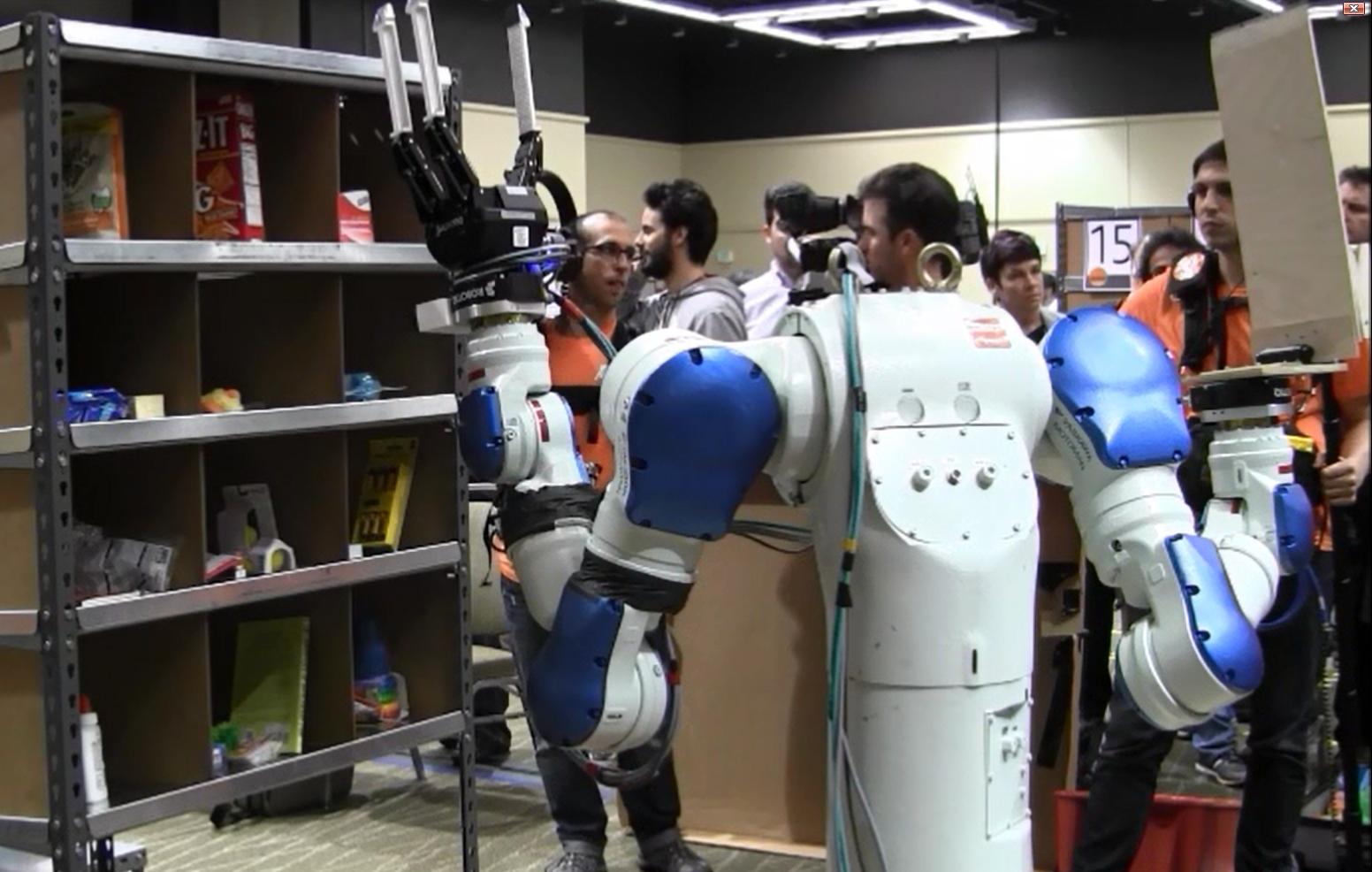}
\caption{Yaskawa Motoman with a Robotiq 3-fingered hand with custom extensions and scooping mechanism used by the Worcester Polytechnic Institute (WPI) team.\label{fig:wpi}}
\end{figure}

All but one team either used a single arm (9) or a multi-arm robot (15), such as the Yaskawa Motoman (Figure \ref{fig:wpi}) or Baxter. Team SFIT opted for a multi-robot solution that involved twelve small differential wheel robots, each equipped with a camera and small gripper, that dragged items out of the bins onto a conveyor belt. 

Six teams opted for a mobile base, whereas two teams employed a gantry system to increase the workspace of their solution. Examples are PickNik who mounted a Kinova arm on a custom gantry, Grizzly which equipped a Baxter robot with a mobile base to be able to pick every bin with either the left or the right arm, using suction on one and a hand on the other, RBO who used a Barrett WAM arm on a mobile base, and Team MIT who used a single arm large enough to reach every bin without additional mobility. 

\begin{figure}
\includegraphics[width=\columnwidth]{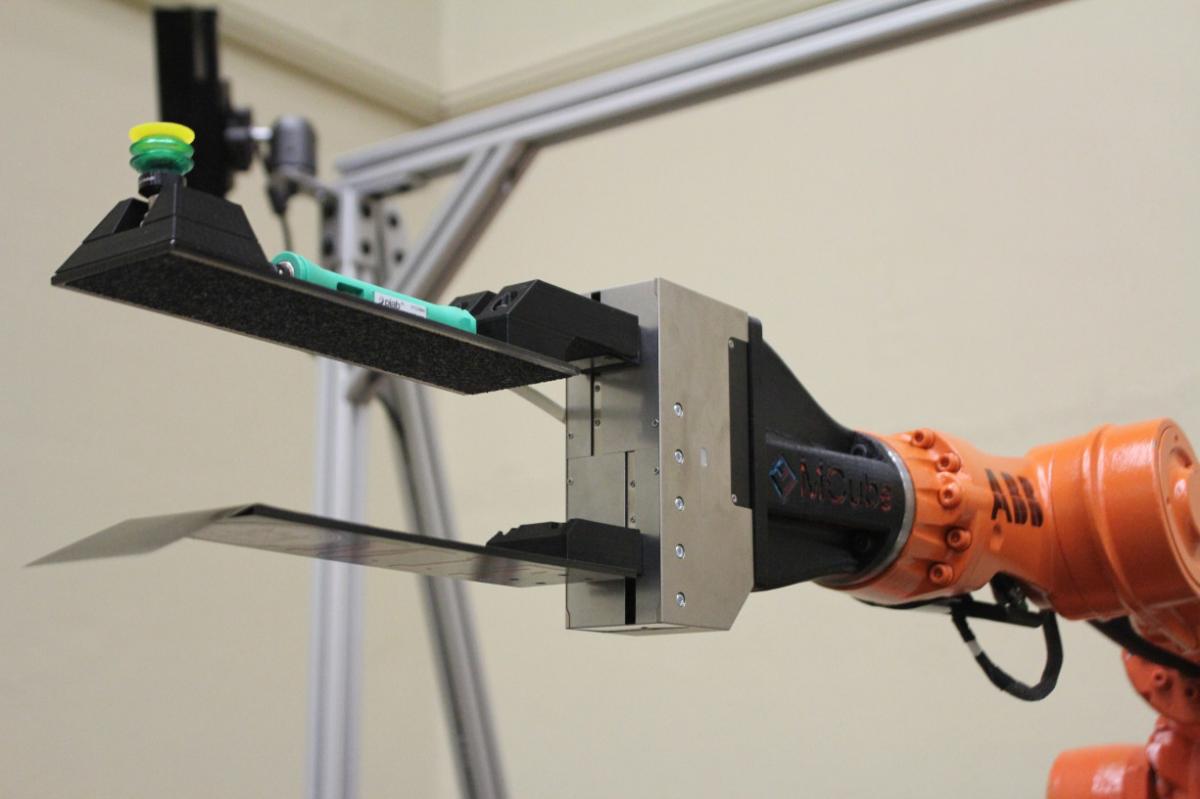}
\caption{Custom gripper by team MIT. At the outer-most end of bottom finger tip is a spatula-like finger nail to scoop objects that are flush against a shelf wall or from underneath. A suction system on the top finger can be employed for items that are hard to grasp.
\label{fig:mithand}}
\end{figure}

For end-effectors, 36\% of the teams (9) used some form of suction, see for example Figure \ref{fig:mithand} showing Team MIT's solution, whereas 84\% (21) teams relied on force-closure and/or friction. That is, only four teams relied exclusively on suction, including the winner of the competition, team RBO, whereas five teams employed a combination of both. Comparing the choice of end-effector to the actual performance, we observe the following:  From the 13 teams who scored better than zero points in the competition, eight teams used some form of suction and only five teams relied exclusively on force-closure and/or friction. These teams were NUS Smarthand (Kinova two-fingered gripper, ranked 4th),  C$^2$M (Mitsubishi gripper, ranked 6th), GeorgiaTech (Schunk gripper, ranked 11th), Duke (Righthand Robotics, ranked 11th) and CVAP (PR2, ranked 13th). Altogether, fourteen teams used off-the-shelf end-effectors, including the PR2 gripper (3), Robotiq gripper (3), Kinova hand (2), and one team each a Baxter gripper, a Barrett hand, the Pisa-IIT Soft hand, a RightHand Robotics Reflex hand, a Schunk hand, or a Weiss parallel jaw-gripper.

\begin{figure}
\includegraphics[width=\columnwidth]{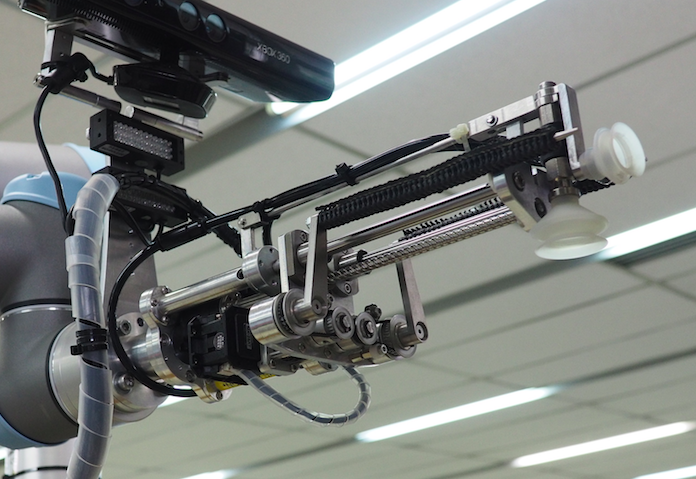}
\caption{Team Nanyang's custom gripper has two suction cups and one pair of fingers. The choice of suction to use depends on the object position. The fingers, which can extend forward and downwards, are used mainly for objects with organic shapes.
\label{fig:nanyanggripper}}
\end{figure}

In some cases, teams combined these grippers with suction---or in the case of Plocka Plocka used the gripper to hold a suction tool on demand---combining the advantages of both approaches. In the ``custom'' category, teams employed various suction systems, often involving off-the-shelf ``contour-adjusting suction cups''.\footnote{Text quotes are taken from the survey.} In this context, ``contour-adjusting'' refers to the fact that the suction cup itself is soft and is therefore able to comply to the object's surface to increase the effectiveness of suction. The MIT team combined suction with a ``spatula-like finger nail'', which allowed the robot to ``scoop objects from underneath, or grasp objects that were flush against a shelf wall'' (Figure \ref{fig:mithand}).  Team Grizzly used a combination of suction and grasping using Baxter's stock suction cap in one hand and the Yale Open Hand \cite{ma2013modular} in the other. The online choice of which tool to use for each target item was based on previous performance data for each method and object. The Rutgers U. Pracsys team collaborated with a company, Unigripper, to design a custom-size vacuum gripper with a wrist-like DoF, which has multiple openings where vacuum is generated. The Unigripper tool for Rutgers was combined with a Robotiq 3-finger hand. Team Nanyang deployed a gripper that combined two suctions and parallel fingers. The dual suction mechanism allows the gripper to suck items from the front, top, or side. The choice of which picking mechanism to use depends on the item pose, item position inside the bin, presence of other items, and previous performance of using the mechanisms with the items. Team CVAP  modified the gripper of a PR2 to be ``thinner'', allowing them more mobility inside a bin. The University of Alberta team combined a Barrett hand with a ``push-pull mechanism'' consisting ``of a flexible metallic tape, step motor and a roll mechanism'', that allowed them to push and pull objects inside the bins.

Asked about how team would change their design, we identified the following recurrent themes in the free-form answers: ``Change gripper supplier/design'' (8), ``Use suction'' (7), ``Making the end-effector smaller/thinner to improve mobility'' (4), ``Increase workspace of the robot / add mobile base'' (4), ``Enhance gripper with sensor/feedback'' (2), ``Complement suction with gripper'' (2). Minor changes include problems with payload restrictions (``gripper too heavy'') and suction systems being too weak. Team SFIT, who employed a team of twelve miniature robots placed on a separate shelf with floor heights identical to that of the shelf in which the items were placed, reported only minor design revisions, including reducing the overall number of robots. 

\begin{table*}[htb!]
\begin{center}
  \begin{tabular}{| l | c | c | c | c | c | r |}
    \cline{2-7}
    \multicolumn{1}{l|}{} & Head & Torso & End-effector & Arm & Mobile base & Total respondents \\ \hline
    3D imaging (Kinect, Asus, etc.) & 8 & 3 & 6 & 6 & 0 & 20 \\ \hline
    2D imaging (camera) & 4 & 1 & 4 & 1 & 0 & 7 \\ \hline
    Laser scanner & 0 & 0 & 1 & 0 & 4 & 5 \\ \hline
    Distance sensor & 1 & 0 & 2 & 0 & 0 & 3 \\ \hline
    3D imaging (tilting laser scanner) & 0 & 2 & 0 & 0 & 0 & 2 \\ \hline
    Tactile sensor & 0 & 0 & 1 & 0 &0 & 1 \\ \hline
\end{tabular}
\end{center}
\caption{``\emph{What kind of sensors did you use and where were they mounted?}''. Cells contain the number of sensors that teams have deployed per location. The column ``total respondents'' contains the number of teams that used each sensor modality. \label{tab:sensors}}
\end{table*}

\begin{figure*}[!htb]
\includegraphics[width=\textwidth]{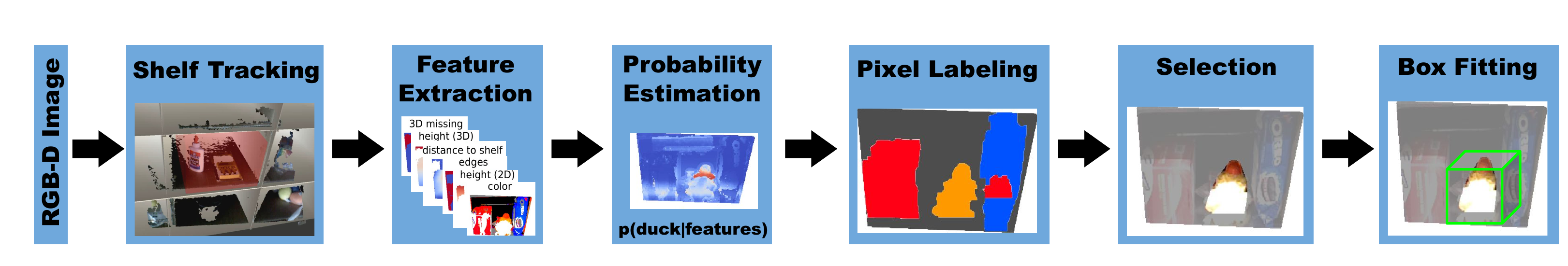}
\caption{Object recognition pipeline employed by the RBO team \cite{eppner2016,Jonschkowski-2016}. \label{fig:rboperception}}
\end{figure*}

\subsection{Perception}

What kind of sensors were used and where they were mounted is summarized in Table \ref{tab:sensors}. 3D sensing in some form was employed by 22~teams, 20 of which using structured light or time-of-flight sensors, such as the Microsoft Kinect, Asus Xtion, or Intel Primesense. These sensors were mounted at various locations on the robots, most often at the end-effector (6), arm (6), and the head (8), but also on the torso (3). Several groups used more than one sensor (20 teams used a total of 23 different sensors.). Conventional 2D imaging was used by seven teams, mounted on the end-effector (4), arm (1), head (4) and torso (1). Only University of Alberta employed a laser scanner on the end-effector, whereas four teams reported using a laser scanner on a mobile base, presumably in support of navigation and alignment with the shelf rather than object detection. Two teams reported using a distance sensor in the end-effector, with one team mounting it at the robot's head (Baxter), albeit it is unclear whether this configuration was relevant for the competition. Only Team RBO reported using a pressure sensor at the end-effector to identify contact, while three teams mention torque/force sensing at the robot's joints for this purpose. 

\begin{figure}[!htb]
\includegraphics[width=\columnwidth]{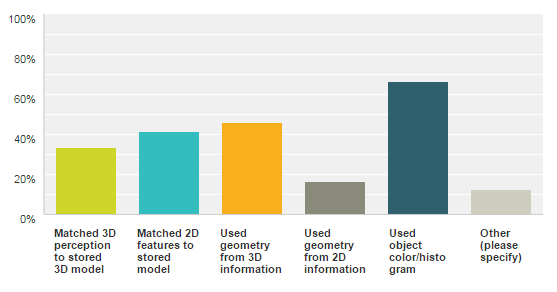}
\caption{``Describe your object recognition approach'' (multiple answers possible).\label{fig:apc_perceptionapproach}}
\end{figure}

\begin{figure}[!htb]
\includegraphics[width=\columnwidth]{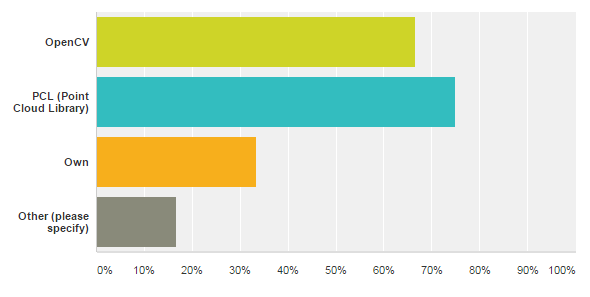}
\caption{``\emph{What software libraries did you use?}'' (multiple answers possible).\label{fig:apc_perceptionsoftware}}
\end{figure}

In terms of perception algorithms, 67\% of the teams (16) reported using object color and histogram data, 46\% (11) used geometrical features from 3D information, 42\% (10) matched image features to those stored in a model, 33\% (8) matched 3D perception data to a stored 3D model, and 17\% (4) used geometrical features from 2D information (Figure \ref{fig:apc_perceptionapproach}). A sample perception pipeline employed by the RBO team is shown in Figure \ref{fig:rboperception}. As for the software used, 75\% (18) of the teams used the ``Point Cloud Library'' (PCL) \cite{rusu20113d}, 67\% (16) used the ``Open Computer Vision Library'' (OpenCV) \cite{bradski2008learning}, and 33\% (8) report using their ``own'' tools (Figure \ref{fig:apc_perceptionsoftware}). Other mentions (once each) include: ``Object Recognition Kitchen'' (ORK) \cite{ork_ros}, ``SDS'' \cite{BharathECCV2014}, ``Linemod'' \cite{linemod}, ``Ecto'' \cite{ecto} and ``Scikit learn'' \cite{pedregosa2011scikit}, as well as proprietary software provided by Capsen Robotics (used by the MIT Team). 

Looking at this data more closely and inspecting verbatim answers by the teams, we observe a large diversity of approaches while also observing some common trends. Four teams indicated they used exclusively color and histogram information to identify objects, two teams only used feature detection, such as SIFT \cite{tuytelaars2008local}, and another two teams relied exclusively on matching 3D perception against 3D models stored in a database \cite{aldoma2012point}. With a third of teams relying on only one class of algorithms, the majority of teams chose more or less complex combinations thereof. Three teams combined color and histogram information with point cloud-based background subtraction exploiting the known shelf geometry. Two teams instead used color information for segmentation and then used the remaining point cloud for pose estimation. Another two teams used the object's known geometry (bounding box size) for classification, with and without exploiting 2D image features. The remaining nine answers each report distinct combinations of color and 3D information including \cite{domae2014fast,DBLP:journals/corr/RennieSBS15}. Team Z.U.N. did not reply to the question. Here, Team-K's approach is noteworthy in that they performed identification only after picking up the object and placed it back into the bin if it was not the desired one. 


Asked what teams would do differently, we identify two clusters of responses: those who would like to complement their approach with the algorithms and sensors they did not use (14), and those who would like to simplify their approach (6). Team Duke would not make any changes and four teams did not respond to the question. Among those who want to make improvements by increasing functionality, using more 3D perception and object geometry (5), using color and histogram information (2), and exploiting texture/features (2), were the common themes. Those that wish to simplify their solutions mention problems with 3rd party software packages and computational cost, which they hope to alleviate by falling back on more standard open-source products (OpenCV/PCL). 

\subsection{Planning and Control}

Almost all of the teams (20) implemented some kind of heuristic that took into account both difficulty of the task (presumably features like the number of items in the bin) and previous experience that allowed the designers to associate different success rates with different objects. Only three teams used simpler algorithms like sequentially moving from bin to bin or picking the object that is closest from the current end-effector position. The only team that did not implement a high-level planning algorithm to address sequencing is team SFIT, which employed twelve robots working in parallel.

\begin{figure}
\includegraphics[width=\columnwidth]{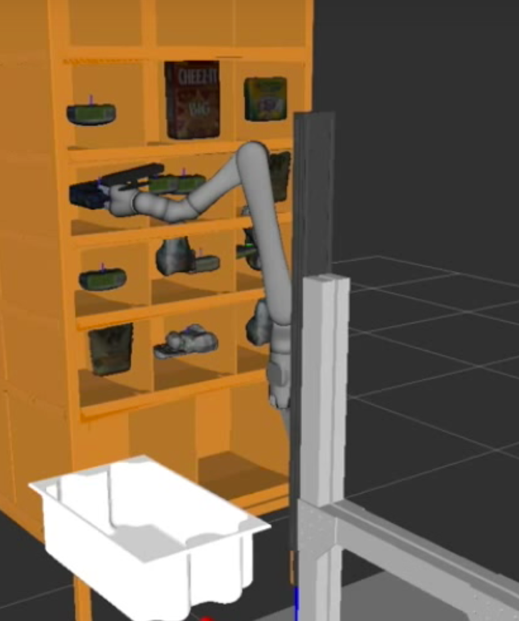}
\caption{Team PickNick's system modeled in the MoveIt! framework \cite{coleman14b} that allows access to large variety of path planning tools and visualization.\label{fig:moveit}}
\end{figure}

We also queried the teams on their use of motion planning software. 80\% of all teams (20) did use motion planning, whereas 20\% of teams (5) did not use any motion planning (i.e., searching for paths in a configuration space representation, as opposed to sensor-driven reactive control). Notably, the winning team (RBO) did not use motion planning. Team MIT relied on trajectory generation using the available software Drake~\cite{drake}, a planning and control toolbox for non-linear systems. The third place Grizzly team used very simple, home-made motion planning to align the robot with the shelf and then servo to pre-computed positions. Among the available motion planning software solutions, ``MoveIt!'' \cite{coleman14b}, Figure \ref{fig:moveit}, was used by 44\% of the teams (11), 28\% (7) developed their own custom solutions, team Georgiatech used ``OpenRave'' \cite{diankov2008openrave}, and team MIT used ``Drake'' \cite{drake}. Other tools mentioned by the teams include ``trajopt'' \cite{schulman2013finding}, ROS' JT Cartesian Controller \cite{nakanishi2005comparative}, and the OMPL library \cite{sucan2012open}, which was interfaced through MoveIt! or stand-alone in the teams' custom implementations. For grasping, 96\% of the teams (23) reported having developed their own, custom solution. The Rutgers U. Pracsys cited using the ``GraspIt'' software package \cite{miller2004graspit} for generating grasping poses for the 3-finger Robotiq hand. 32\% of the teams did use a dynamic IK solver, 60\% did not, and 8\% (2) were not sure. This is important as this capability is often part of the lower-level control software, such as in the Baxter robot, which is not accessible to the designer. Software packages used by the teams include the ``Rigid Body Dynamics Library'' (RBDL) \cite{rbdl}, OROCOS with its ``Kinematics Dynamics Library'' (KDL) \cite{bruyninckx2001open}, ``Drake'' \cite{drake}, ``EusLisp''\cite{Matsui90EusLisp} and ``Klamp't'' \cite{klampt}.

``Visual servoing'' was used by only 8\% (2) of the teams. The other 92\% (22) indicate they did not rely on visual servoing. ``Force control'' was used by 20\% (5) of the teams, whereas 80\% (20) ignored the forces induced in the robot during task execution. 

When asked what they would change, introducing more reactive control was the dominant response (8) from 22 teams responding to this question. This entails adding feedback that helps the robot ascertain that it really holds the object, as well as using force feedback and visual servoing to make up for uncertainty in sensing and actuation. Four teams indicated that they wish to simplify their motion planning approach to have more direct access to path planning than MoveIt! provides. Another four teams wish to improve grasping by better training grasp approaches for known objects, but also investigating techniques that exploit the environment, e.g., by pushing an object against the wall. Two teams would like to improve their robot's model fidelity and how to define tasks and constraints in this space. Team Grizzly indicated that using a more established architecture, such as ROS, would be desirable to abstract more complex actions such as controlling two arms at once, and team Robological wishes to use formal methods for controller synthesis and validation. The remaining replies express the desire to have done more training and fine-tuning.  

\subsection{Summary questions}

\begin{figure}[!htb]
\includegraphics[width=\columnwidth]{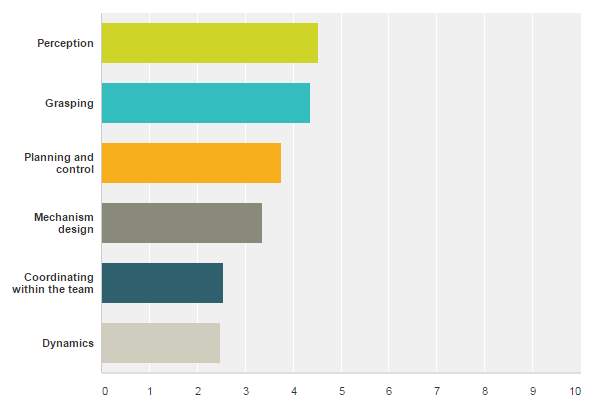}
\caption{``Please rank order the different aspects of the APC by their difficulty, starting with 'most difficult'  at the top''.\label{fig:apc_challenges}}
\end{figure}

We asked a couple of summary questions. First, we asked the teams to rank-order the challenges they encountered, letting them chose from six categories: ``Perception'' (4.52), ``Grasping'' (4.36), ``Planning and Control'' (3.75), ``Mechanism Design'' (3.36), ``Coordinating within the team'' (2.54) and ``Dynamics'' (2.48). The number in parenthesis is the average score, where higher scores correspond to being ranked ``harder''. This data is shown in Figure \ref{fig:apc_challenges}. Looking more closely at the data, most teams ranked the categories in a similar order (similar distributions around the mean score) with the exception of team coordination, which was the biggest challenge for four teams, and the least for eleven. 

We also tried to gauge the teams' opinions on a series of general questions. 84\% (21) of the teams either strongly (11) or somewhat agree (10) to the statement ``\emph{Perception needs to be better integrated with motion planning.}'' Two (2) teams are neutral, and two (2) teams ``somewhat disagree''. 68\% (17) of all teams either strongly (10) or somewhat  (7) agree to the statement ``\emph{Motion planning needs to be better integrated with reactive planning.}'' Six (6) teams are neutral on this statement, and two (2) do not agree. Finally, 60\% of all teams either strongly (8) or somewhat (7) agree to the statement ``\emph{Development of capable, human-like robotic hands is not on the critical path for widely deploying autonomous robots}''. Five teams are neutral on this statement and another five ``somewhat disagree''.  

\section{Analysis of survey results}\label{sec:analysis}
We will now provide an analysis of the survey results presented in Section \ref{sec:results}. While some of our insights are based directly on the data, both in terms of statistic and comments by individual teams, others are based on the personal opinion of the authors, who are representative only for a subset of the participants and organizers. 

\subsection{Team composition}

Participants were, to a large part, graduate students, post-docs or other professionals (81\%). Given that 24 of the 25 teams were from an academic environment, the high number of post-graduate participants is unusual for a robotic competition---a format that is popular in senior undergraduate robotics classes~\cite{almeida2000mobile, messom2002robotic, tougaw2005integrating, chew2009robotics, pastor2008participating, nagatani2012sensor, correll2012one}. We believe this to be due to the high complexity of the task, which involves mechanical design, perception, planning and grasping. It is possible that packaging a bare bones version of the competition could create a framework for teaching a class around the Amazon Picking Challenge.  Recent experience from Rutgers University indicates that students appreciate a semester-long project around the challenge. For this, the problem complexity must be reduced (easier access to objects, fewer object categories) and  access to existing software and hardware solutions (e.g., Baxter, OpenCV, etc.) should be provided. Defining such an easily accessible framework and testing it across different institutions may be very valuable to the community.

We also observe that involvement from the commercial sector was minimal (three out of 25 teams). Team Robological identified as a start-up of which some members are still affiliated with the University of Sydney, whereas Team Grizzly and  C$^2$M are affiliated with Dataspeed Inc.\ and Mitsubishi Electric Corp., respectively, which are established companies in robotics and automation. We note that none of the competitors identified as exclusively commercial, with team Grizzly involving students from Oakland University, and  C$^2$M students from Chubu University and Chukyo University. We believe, due to lack of data from the industrial sector, that the requirement to release and allow open-source access to all software packages and mechanical designs was a deterrent to commercial labs. That being said, interest from robot manufacturers was high, with multiple companies lending hardware free of charge and/or providing it to teams to use at the competition site (Barrett Technologies, Clearpath Robotics, Fanuc, Rethink Robotics, Universal Robotics, Yaskawa Motoman, UniGripper, Robotiq, ABB). 

\subsection{Mechanical design}

Although the design approaches varied widely, from large, static single robot arms to mobile two-arm manipulators, it is difficult to identify a platform that is ``best''. Although the challenge made use of only the middle section of bins, this was still at the working limit of many of the commercial robots. Among the top three teams, two employed a mobile platform and the other employed a single large arm.  Secondary metrics, such as overall space consumption, power, or the ability to deliver objects elsewhere were not challenged by this competition. In the long run, speed will be a significant factor in many industrial applications, which may give an advantage to static arms and gantry solutions over wheeled platforms. While there was a significant number of teams in the first APC that used dual-arm manipulators, there were few attempts to pick items in parallel with the two arms, which is one way that faster picking can be achieved. Moreover, the advantages of dual-arm robots may be potentially more significant in future competitions with increased number of objects. In cases of significant occlusions, one arm can be used to clear a blocking object while another attempts to grasp the target object. 

Regarding end-effector design, there is a clear trend in support of suction-based approaches. Suction alone (RBO, Team-K, Team A.R.) was proven sufficient and there were solutions that aimed to combine suction and friction-based grasping in this competition, including the two runner ups Team MIT and Grizzly. Creating constraints by grasping requires careful alignment of opposing forces on the object, whereas sucking requires only a single area of contact with the suction orifice. Unlike friction-based grasping, sucking an object minimizes both translational and rotational degrees of freedom, which makes the approach robust against wrenching forces. Many teams with a vacuum-based approach used an off-the-shelf vacuum cleaner or suction cup, instead of industrial setups. Some traditional suction mechanisms used in the industry can require careful placement to maintain vacuum, whereas vacuum cleaners continuously pull the air, which may work even if there is an opening between a gripper and an object, due to a complex surface or unexpected motion of the arms. This more robust attachment comes at the cost of much longer decay times when releasing an object.   A drawback of suction-only approaches are their poor ability to manipulate an object; this capability was not prominent in the APC so far, however future contests are expected to have more populated bins that may require more flexible manipulation and object rearrangement \cite{krontiris2015rearrangement}, or a more explicit exploitation of the environment \cite{ChavanDafle2014,eppner_ijrr_2015}. The value of a combination of suction and grasping also becomes clear when considering objects like the metal-mesh pencil holder or the foam balls, which are difficult to suck. With only two out of 25 items having this property, however, this first competition encouraged using a less complex suction-only solution than dealing with the challenges that a combined approach entails.  

Most teams would re-design or improve their grasping capabilities. Half of the 14 teams that did not use suction would change their design to include suction. Almost a third of the teams (8), would either change the gripper they were using or dramatically improve it to become more dexterous, thinner and more light-weight. 

The most unique mechanical design, Team SFIT, involved twelve miniature mobile robots. This approach was among the many teams that did not successfully score any points, which makes it difficult to compare its merits against other designs quantitatively. Philosophically, however, their unconventional approach could offer a variety of benefits. Mobile robots that are individually smaller than most grippers used in the competition could possibly reach far into the corners of each bin, while their number increases robustness to mechanical breakdown. On the other hand, smaller robots may not have sufficient strength to extract heavy items or the flexibility to deal with stacked or occluded items. Furthermore, an increased number of mechanical components may decrease the robustness of the overall solution.

\subsection{Perception}

With 20 teams employing structured light for 3D perception, this technology was by far the most used sensing modality. Although perception turned out to be a key challenge in this competition—--with many groups working around the grasping problem by employing suction—--it is not easy to identify a correlation between the teams' background in perception and their performance in the competition. Indeed, all top three teams identify vision as one of their key challenges and note insufficient background in their groups. At the same time, groups with a known track-record in computer and machine vision were not as successful. A reason for this might be the maturity of open source tools such as PCL and OpenCV, which allowed most groups to cover their basic sensing needs quickly.  Another explanation could be that the perception problem in the APC actually differed from known vision problems; this would be supported by the fact that the winning team TUB developed a perception module tailored to the challenge~\cite{Jonschkowski-2016}. Overall, research-grade software provided often only marginal improvements while it lacked the maturity of well-maintained open-source projects. In future competitions, as the item packing density increases in the bins, more advanced vision software may play a more differentiating role. 

To assist the development of visual perception solutions for solving warehouse pick-and-place tasks, a new rich data set has become available that is devoted to this type of challenges~\cite{DBLP:journals/corr/RennieSBS15}. The publicly available Rutgers data set includes thousands of RGBD images and corresponding ground truth data for 3D object poses for the items used during the first Amazon Picking Challenge at different poses and clutter conditions. 

In the survey, only a few teams expressed a desire to enhance their gripper with sensors. This is surprising, as only three of the presented solutions actually had any sort of feedback in their grippers, and only two used it. It would appear that adding sensing to the end-effector can improve robustness: Employing a pressure sensor was integral to RBO and MIT team's suction capabilities, and Team-K read the control board of the vacuum cleaner to detect the sucking status, whereas Duke did not use the sensors provided by the RightHand Robotics ReFlex hand. Team MIT used force feedback in the opening of their parallel-jaw gripper to detect contact with the shelf and the objects. Other teams used visual and force sensing to detect whether a grasp was actually successful or to re-adjust grasping. One reason for this might be that the community as a whole has very little experience with in-hand sensing due to the lack of availability of hands with integrated sensors and algorithms that use this information during grasping. Indeed, there are only a small number of such systems out there, and only few are commercially available \cite{tenzer2014feel}. 

\subsection{Planning and Control}

Most teams employed a high-level task planning framework that targeted maximizing the expected score, many of which went for difficult objects that were likely to lead to penalties for dropping or picking the wrong item. While this strategy was appropriate for a game, a production setting would require all items to be picked. For such an environment, it would be more interesting to find policies that maximize throughput by minimizing trajectories (which only Team Nanyang did) or exploiting the ordering of items in a bin. 

Surprisingly, many of the teams, including the winning team, did not make use of motion planning. Here ``motion planning'' refers to a deliberative algorithm that uses environment and robot models to generate a collision-free trajectory before executing it. It was possible to build successful systems without motion planning because the real-world scenario on which the competition was based was designed for easy picking by humans. Easy access to all of the bins and easy access to the objects within each bin effectively eliminated the need for complex motion planning around obstacles. As a result, reactive control approaches were sufficient to generate appropriate motions while avoiding obstacles within the shelf.  

While a large number of teams used MoveIt!, an integrated motion planning and visualization framework, none of the top three performers used such software. This may suggest---like in the case of perception software---that prepackaged toolkits for these complex behaviors help teams to get started rapidly \cite{coleman14b}, but do not necessarily help them access and improve lower-level functionality in an equally easy manner. Generally, MoveIt! and other pre-packaged motion planning software solutions have three problems: 1) the robot is not allowed to exploit contact, 2) uncertainty is not taken into account during planning, and 3) incorporation of sensor-based feedback is not straightforward. This approach is in contrast with the winning team's architecture that consisted of a hybrid automaton that connected a variety of feedback controllers \cite{khatib1987unified} with event-based state transitions. Here, sensors included object position provided by the camera, contact via pressure sensors, and actual torques. Since motion planning appears important in general in geometrically more complex scenes to navigate around obstacles, a potentially important topic of future research is how to better integrate planning with feedback to make up for inaccurate sensing and actuation. There currently exist no high-level tools that combine these approaches in a user-friendly way. The community would greatly benefit from manipulation planning tools that better support reasoning over contacts, sensor-based feedback and uncertainty. 

Regarding grasp planning, an overwhelming majority of teams used custom approaches, which may appear surprising at first. To explain this, we first note that about 20\% of teams opted for suction over grasping, which dramatically simplifies the problem by reducing it to choosing surfaces that are flat and planning to reach them. However, this level of customization is also indicative of more fundamental problems, namely, difficulties in generalizing the grasping problem across mechanical platforms, and difficulties in incorporating uncertainty and environmental context into grasp planning.  The output of ``GraspIt!'' designates an end-effector and finger pose for a given object geometry that optimizes some wrench-based grasp metric.  This metric ignores environmental context, reachability, pose uncertainty, and nonprehensile strategies, such as pushing, that may be more important than robustness to disturbance wrenches. Further tipping the balance toward custom solutions is that current trends in manipulation include shifting some of the required reasoning into end-effector compliance \cite{dollar2010contact, ma2013modular} and using under-actuated systems \cite{kragten2010ability, Mason2012, deimel2014novel, eppner_ijrr_2015}. As a result, many objects can be grasped using simple rules, such as attempting a power grasp along the medial axis of the object.  
Compliance can play an important role even in conjuction with suction as indicated by the Unigripper's design with the Rutgers U. Pracsys team, where a foam is introduced between the object and the suction openings so as to help to adapting to the surface of an object and forming vacuum by pressing on the object.

It is important to note that the APC shelves are relatively uncluttered compared to the shelves encountered by human pickers, which may have dozens of objects in close contact.  This simplification may have biased the teams' choices of grasping strategies toward solutions like suction and standard parallel-jaw grippers, whereas a more complex arrangement of objects may motivate the use of human-like dexterity and grasp planning capabilities.

There appears to be a division between control-centric and planning-centric approaches to problems like the APC. That is, some teams relied exclusively on control-based approach (visual servoing and force control), not using any planning, whereas others relied exclusively on planning. 
Which approach is better can unfortunately not be concluded from the teams' performance. The two top solutions made extensive use of visual servoing and force control.  They also did not involve ``grasping'', i.e., explicit reasoning regarding grasping poses, and only a very limited amount of ''motion planning'', i.e., collision-free planning in a configuration space representation.  In contrast, the third-place team (Grizzly) did not use any reactive control schemes, but relied on motion planning and SLAM (Hector SLAM ROS package \cite{KohlbrecherMeyerStrykKlingaufFlexibleSlamSystem2011}) to localize and move a mobile Baxter robot in front of the appropriate bin. It is therefore unclear what the ``best'' approach is, albeit the strategy of using reactive control to compensate for inaccurate sensing and actuation of an underlying deliberative architecture appears to be powerful. 

Indeed, when asked what to improve, there was a clear desire to include more reactive control to make up for deficiencies with the sense-plan-act model. The challenge participants were also not content with the abstraction level that prepackaged software solutions like MoveIt! provided. On the one hand, teams wished for the ability to model their robot hardware more easily and have simple ways to provide tasks and constraints, much like the way MoveIt! \cite{coleman14b} and OpenRave \cite{diankov2008openrave} provide. On the other hand, the tools are not perfect yet, are difficult to debug, and have a high learning curve should a solution require the team to make changes ``under the hood'' of such tools. A possible solution here might be not only to continue to improve these tools, but abstract their lower-level functionality into a higher level language, making their inner workings more accessible, and making it easier to attach arbitrary sensing, reactive controllers and logic to the trajectories they generate. 

\section{Discussion}\label{sec:discussion}

An important conclusion to draw from the APC is that recent developments in robotics have the potential of substantially increasing the degree of automation in warehouse logistics and order fulfillment in the near future.  Many efforts to broaden the impact and applicability of robotics in industry beyond factory automation have faced substantial challenges.  The kind of warehouse logistics addressed in the APC, however, can believably be automated using existing or near-future technologies and potentially faster than many other target applications of robotics.  It therefore seems worthwhile to continue the APC in order to foster the exchange between the robotics community and relevant industrial partners. 

Addressing warehouse logistics and order fulfillment in industrial settings will probably still require substantial scientific progress. As was outlined above, some of the standard solutions, such as motion planning or complex hands, were not necessary to succeed in the first instantiation of the APC. This may point to the fact that the space of possible solutions is not fully explored yet and that simple approaches may be a more promising route for critical applications despite the importance of providing general-purpose robots.  It is possible that the focus on component technologies, such as 3D object pose estimation, control, motion planning, grasping, etc., has not allowed the community to study integrated solutions.

The challenge of system integration is demonstrated by the fact that half of the teams did not score any points despite having developed impressive sub-systems. For example, Team A.R.\ looked very promising in warm-ups, but the particular product arrangement they drew for the trial had the glue bottle alone in a particular configuration in the lower left bin. Their system's planner generated a path through a configuration that required rotating the end-effector in such a way that the vacuum hose wound around the arm. The team had not adequately considered the hose behavior in their planner, and this particular situation exposed a corner case they had not seen during development and testing. Other teams failed because of last minute software changes, or failures to model the lip of the shelf such that the gripper had trouble finding a way into the bins. Lighting in the convention hall also proved to be a problem for some teams.  For example, the Duke team resorted to taping an umbrella to the top of their robot to block overhead light.

A more comprehensive treatment of robotic challenges, in terms of their software, hardware and algorithmic components, appears necessary. In particular, in this competition teams were faced simultaneously with a hardware and a software design problem. This allowed to simplify the complexity of the software development process by modifying hardware, or vice versa. For example, the use of vacuum grippers side-stepped the challenging problems of grasp planning and in-hand manipulation, which are more critical when using human-like hands. Thinner end-effectors simplified the process of computing collision-free paths in tight spaces. Integrating sensor-feedback in the control process was used successfully by several teams to compensate for less precise actuators, such as mobile bases or lower-cost robot arms. This pattern highlights the continued need for cross-disciplinary collaboration in robotics between hardware, software and algorithmic researchers to build task-specific, robust integrated systems.

The APC showed once again that system integration and development remain fundamental challenges in robotics. When a working system consists of dozens, if not hundreds, of independent components and the failure of each of these components can lead to catastrophic failure of the overall system---as witnessed during the competition---the focus is shifted from scientific questions towards software and hardware development and testing practices.  As a community, it is critical to decide whether such insights should be equally worthy of publication as technical advances.  It also suggests a need for a common and accepted knowledge-base of how to build, test, and deploy integrated robotic solutions.  It is arguable whether this expertise already exists in industry, where the complexity of an application is most frequently addressed by structuring the environment or tailoring parts for robot handling. These approaches do not extend to the type of problems addressed in robotics research. 

\subsection{Moving Forward}

While the first APC laid a foundation for testing competing solutions to order fulfillment, the manipulation problem was greatly simplified relative to real-world warehouse scenarios. In particular, we discuss three axes in which the complexity of APC can gradually increase to get closer to a real scenario.

\myparagraph{Object Packing Density.}
The cost of land and indoor spaces stresses the need for packing more items into smaller spaces, shelves and bins. This creates the need for picking, placing, and manipulation in tight spaces and for tightly arranged objects. 

In the APC 2015, the objects in the bins were arranged side by side and lightly packed, far from what would be expected in a warehouse. Tight object arrangement has important consequences for the manipulation strategies employed, and for how the robotic system interacts with objects and storing structures. Extracting a free-standing book from a bin and extracting a book that is wedged between other books are very different manipulation problems. While the first can be solved in the pick-and-place paradigm (i.e., reach, grasp, extract), the second is badly suited to standard grasp planning techniques. The desired contacts surfaces are rarely sufficiently exposed, leading to a different manipulation problem where the grasp is only the last stage of a longer process that drives the object into the gripper and where interactions with the environment play a critical role.

Sparse bins allow methods that avoid contact with other obstacles to be successful. In APC 2015, once an object was grasped (whether by suction or by fingers), it could be directly extracted. However, more tightly-packed bins may require brushing obstacles aside in order to reach and retrieve the target object, as well as sliding the object along the bottom or sides of the bin. The need to manipulate in contact may radically change both the software and hardware methods used in the APC (e.g., shifting the emphasis toward compliant control and compliant hardware), and the level of sensing required at the point of the manipulation (e.g., tactile and in-hand vision sensors). A key challenge to overcome in tightly-packed bins is to perform the necessary manipulation under limited sensing and poor prior knowledge of the environment.

\myparagraph{Speed.} Human pickers in Amazon warehouses pick items at an approximate rate of 5-10 seconds per item. Reaching that speed with an automated solution is likely as much a research problem as it is an engineering one, requiring fine-tuning computations of all algorithms as well as optimizing all robot motions. It is therefore not a reasonable goal for APC to expect that average rate. Improvements in speed are nevertheless a direction in which the challenge could propel us forward, if these technologies are to become useful in the near future. With the danger of leading the community to premature optimization rather than out-of-the-box innovation, speed can be used as one measure of progress, potentially guiding the selection of robotic mechanisms as well as algorithmic solutions.

\myparagraph{Reliability.} Finally reliability is key to any industrial operation, and the error rates showed even by the top teams were far from the expectations of automation companies. Errors such as dropped items, destroyed items, or miss-classified items should continue to be penalized. 
An interesting variation would be to also include error detection and identification.
That is an type of error that, under the right circumstances might be acceptable, and could have a smaller penalty. 

Tolerance to miss-calibrations plays a key role in the reliability of a system.
During APC 2015 it was possible to off-line and accurately calibrate the relative location of the shelf with respect to a static robot. This simplifies manipulation and is not representative of the real problem. Promoting solutions that are more robust to calibration errors should allow progress towards more flexible systems that can be used in less structured environments. These include warehouses, but also other applications such as home-assistive robots, which will frequently have to deal with manipulation in tight spaces.

\vspace{0.25in}

In recent years, it has been said that grasping is a solved problem.\footnote{Statement from keynote presentation by Gill Pratt at IROS 2012.} That is in part due to a bias toward ``table-top'' manipulation, the DARPA ARM Project being a prominent example. Scenarios with isolated objects without many environmental constraints lend themselves to the grasp-planning approach. APC points to a different problem, one where the key role is not played by the grasp but by the reach and retrieve actions.

While many robotics researchers participated in the APC with great enthusiasm and obtained in return significant insights and advances, there were also critical voices in the community. Several researchers raised the question whether it is appropriate for a technology-oriented company with significant resources, such as Amazon, to divert the work of publicly-financed research labs towards a research agenda beneficial to the company, while investing a disproportionately small amount. Similar arguments were made with the DARPA Robotics Challenge and are probably inherent to the idea of funded challenges.

\section{Conclusion}\label{sec:conclusion}

The APC contest was an exciting showcase of the application of advanced research to a real-world problem. Modern advances in the robotics field are opening up a new set of tasks that are far more nuanced and dynamic than the rote industrial applications of the past. However, it is clear that improvements and breakthroughs are still required to reach human-like levels of speed and reliability in such settings. A human is capable of performing a  more complex version of the same task at a rate of $\sim$400 sorts/hour with minimal errors, while the best robot in the APC achieved a rate of $\sim$30 sorts/hour with a 16\% failure rate. The challenge was an interesting measuring stick that illustrated the maturity of the various components and their readiness to transition into industrial applications.

It is a credit to the robotics community that many of the open-source projects from the robotics world made up the foundation of the APC systems. Developing a system capable of handling such a challenge in a matter of months would be unthinkable without quality tools and applicable research. It is important to note, however, the valuable feedback in places where these tools were difficult to integrate into a full solution, or proved challenging to modify to provide a robust solution to specific tasks.

\section*{Acknowledgements}

We wish to thank all the teams devoting significant resources and time to participate in the picking challenge and providing us with their feedback. 

N.\ Correll was supported by a NASA Early Career Faculty Fellowship. We are grateful for this support. 

O.\ Brock was supported by the European Commission (SOMA project, H2020-ICT-645599), the German Research Foundation (Exploration Challenge, BR 2248/3-1), and the Alexander von Humboldt foundation through an Alexander-von-Humboldt professorship (funded by the German Federal Ministry of Education and Research).

K.\ E.\ Bekris was supported by NSF award IIS-1451737 and the Rutgers U. Pracsys team is grateful for the hardware support of Yaskawa Motoman, Unigripper Technologies and Robotiq.

A.\ Rodriguez and the MIT Team is grateful for the hardware and technical support of ABB.

\appendix

\section{Survey questions}\label{sec:fullsurvey}
We provide here the complete set of survey questions. Answers consisted of multiple choice tick boxes ($\Square$), which allowed multiple answers, radio buttons ($\Circle$), which allowed only one answer choice, and free text comments, indicated by \emph{comment box}. 
\subsection{Introduction}
Thank you for participating in our survey. Our goal is to get a comprehensive picture of the technical approach you have been using, open source tools that you deem most important, where you had to spent most of your development efforts, and what you think the biggest challenges have been. We will compile this information into a comprehensive report that will be made available to the public, providing both important statistics, e.g. on what kind of sensors, hardware or open-source tools you used, and (anonymous) anecdotal trends that we could extract from your written responses. This survey will take 10-15 minutes of your time. The survey consists of seven pages:

\begin{enumerate}
\item This introduction
\item Information about your team
\item Information about the mechanism you constructed
\item Information about sensors and perception algorithms you employed
\item Information about planning and control you developed
\item A set of summary questions
\end{enumerate}

Capturing the complexity of your design is probably impossible using a simple survey. We therefore provide plenty of opportunities to provide answers in text-form, which we encourage you to take advantage of. In order to make this process as smooth as possible, feel free to skip any question you are not comfortable answering. 

\subsection{Information about your team}
We would like to learn more about your team, its background, skill set, and what you think has been missing. 

\begin{enumerate}
\item Your team name. We will use this information to fuse answers from multiple team members to get the most complete picture of your approach and input (\emph{comment box}). 
\item Does your team have an academic or commercial background? (Please check all that apply.)
\begin{itemize} 
\item[\Square] Commercial
\item[\Square] Academic
\item[\Square] Private Party
\item[\Square] Other (\emph{comment box})
\end{itemize}
\item Briefly describe your team including background of your team members (\emph{comment box}).
\item Which skill(s) have been missing from your team? (Please think about technical skills that have not been represented at all within your team, rather than technical aspects that did not work as planned.) (\emph{comment box})
\item Are you the team lead?
\begin{itemize}
\item[\Circle] Yes
\item[\Circle] No
\end{itemize}
\item Please provide us with some performance data (if you remember)
\begin{itemize}
\item Number of items successfully delivered (\emph{comment box})
\item Number of items picked, but lost during manipulation (\emph{comment box})
\item Number of wrong items picked (\emph{comment box})
\item Final Score (\emph{comment box})
\end{itemize}
\item How was your team composed?
\begin{itemize}
\item Professional engineers (\emph{Number})
\item Graduate students (\emph{Number})
\item Undergraduate students (\emph{Number})
\item Other (\emph{Number})
\end{itemize}
\end{enumerate}

\subsection{Your mechanism}
We want to learn more about the mechanism you have designed. You will have a chance to describe perception and algorithmic design choices on the next few pages.
\begin{enumerate}
\item What kind of platform did you use? Please tick all the components you have been relying on.
\begin{itemize}
\item[\Square] Single arm
\item[\Square] Multi-arm robot
\item[\Square] Mobile base
\item[\Square] Gantry system
\end{itemize}
\item What kind of end-effector did you use?
\begin{itemize}
\item[\Square] Suction
\item[\Square] Force-closure/friction
\item[\Square] Electrostatic
\end{itemize}
\item In case you selected ``force-closure/friction'' above, which hand/gripper design did you choose?
\begin{itemize}
\item[\Square] None
\item[\Square] Baxter
\item[\Square] Custom
\item[\Square] Kinetiq
\item[\Square] RightHand Robotics
\end{itemize}
\item Please describe your robot in 3-4 sentences, including kinematics of your platform and end-effector design. (\emph{comment box})
\item How would you change your design? (\emph{comment box})
\end{enumerate}

\subsection{Perception}
We would like to learn about your perception approach and specific tools and hardware you have been using. 
\begin{enumerate}
\item
What kind of sensors did you use and where were they mounted?\\
{\tiny
\begin{tabular}{lccccc}
 & Head & Torso & End-effector & Arm & Mobile base\\
2D imaging (camera) & \Square & \Square & \Square & \Square & \Square \\
3D imaging (MS Kinect, Asus etc.) & \Square & \Square & \Square & \Square & \Square \\
3D imaging (tilting laser scanner) & \Square & \Square & \Square & \Square & \Square \\
3D imaging (stereo vision) & \Square & \Square & \Square & \Square & \Square \\
Laser scanner & \Square & \Square & \Square & \Square & \Square \\
Distance sensor & \Square & \Square & \Square & \Square & \Square \\
Tactile sensor & \Square & \Square & \Square & \Square & \Square \\
\end{tabular}}\\

\item Describe your object recognition approach
\begin{enumerate}
\item[\Square] Matched 3D perception to stored 3D model
\item[\Square] Matched image features to stored model
\item[\Square] Used geometrical features from 3D information
\item[\Square] Used geometrical features from 2D information
\item[\Square] Used object color / histogram data
\item[\Square] Other (please describe) (\emph{comment box})
\end{enumerate}
\item What software libraries did you use?
\begin{itemize}
\item[\Square] OpenCV
\item[\Square] PCL (Point Cloud Library)
\item[\Square] Own
\item[\Square] Other (please specify) (\emph{comment box})
\end{itemize}
\item Please describe your perception approach in 3--4 sentences (\emph{comment box})
\item Perception: What would you do differently?
\end{enumerate}
\subsection{Planning and Control}
We would now like to learn about your solutions to planning and control including the tools you have been using. 
\begin{enumerate}
\item What was your basic strategy for selecting the order in which to pick the items? (\emph{commment box})
\item Did your approach rely on ``motion planning''?
\begin{itemize}
\item[\Square] No
\item[\Square] Yes (OMPL)
\item[\Square] Yes (OpenRave)
\item[\Square] Yes (MoveIt!)
\item[\Square] Yes (Own)
\item[\Square] Other (please specify) (\emph{comment box})
\end{itemize}
\item Did your approach use ``grasp planning''?
\begin{itemize}
\item[\Square] GraspIt
\item[\Square] Own
\item[\Square] Other (please specify) (\emph{comment box})
\end{itemize}
\item Did you rely on a dynamic IK solver?
\begin{itemize}
\item[\Circle] Yes
\item[\Circle] No
\item[\Circle] Don't know
\item[] Please specify the tool you used if applicable (\emph{comment box})
\end{itemize}
\item Did your approach rely on visual servoing?
\begin{itemize}
\item[\Square] No
\item[\Square] Yes (Visp)
\item[\Square] Yes (Own)
\item[\Square] Other (please specify) (\emph{comment box})
\end{itemize}
\item Did your approach rely on force control?
\begin{itemize}
\item[\Circle] Yes
\item[\Circle] No
\item[] Please provide additional information if ``yes'' (\emph{comment box})
\end{itemize}
\item Please describe your overall planning and control approach in 3--4 sentences (\emph{comment box})
\item Planning and control: What would you do differently? (\emph{comment box})
\end{enumerate}
\subsection{Summary questions}
\begin{enumerate}
\item Please rank order the different aspects of the picking challenge by their difficulty, starting with ``most difficult'' at the top
\begin{itemize}
\item[$\updownarrow$] Coordinating within the team
\item[$\updownarrow$] Mechanism design
\item[$\updownarrow$] Perception
\item[$\updownarrow$] Planning and control
\item[$\updownarrow$] Dynamics
\item[$\updownarrow$] Grasping
\end{itemize}
\item What do you think about the following statements?
{\tiny
\begin{tabular}{p{2cm}p{0.7cm}p{0.7cm}p{0.7cm}p{0.7cm}p{0.7cm}}
 & Do not agree & Somewhat disagree & Neutral & Somewhat agree & Strongly agree \\
 Motion planning needs to be better integrated with reactive planning & \Square & \Square & \Square & \Square & \Square \\
 Perception needs to be better integrated with motion planning & \Square & \Square & \Square & \Square & \Square \\
 Development of capable, human-like robotic hands is not on the critical path for widely deployment of autonomous robots & \Square & \Square & \Square & \Square & \Square \\
\end{tabular}}
\item Please provide us with any additional comments you might have? (\emph{comment box})
\end{enumerate}
\bibliography{refs}
\bibliographystyle{ieeetr}

\begin{IEEEbiography}[{\includegraphics[width=1in]{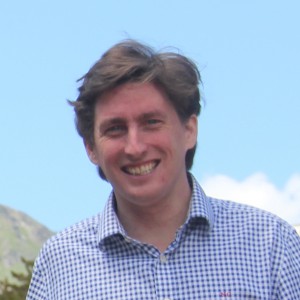}}]{Nikolaus Correll}
is an Assistant Professor in Computer Science at the University of Colorado at Boulder. He received his PhD in Computer Science from \'{E}cole Polytechnique F\'{e}d\'{e}rale de Lausanne (EPFL) in 2007, and a Masters in Electrical Engineering from the Eidgen\"{o}ssische Technische Hochschule (ETH) Zurich in 2003. He worked as a post-doctoral associate at MIT's Computer Science and Artificial Intelligence Laboratory from 2007 to 2009. His research interests include distributed robotic systems, robotic materials and swarm intelligence. He is the recipient of a 2012 NSF CAREER fellowship and a 2012 NASA Early Career Faculty Fellowship, and Best Paper Awards at the International Symposium on Distributed Autonomous Robotic Systems (DARS) in 2006 and 2014. 
\end{IEEEbiography}

\begin{IEEEbiography}[{\includegraphics[width=1in]{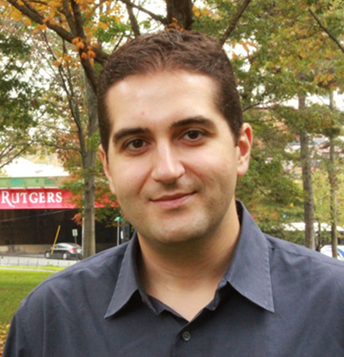}}]{Kostas Bekris}
is an Assistant Professor in the Computer Science department of Rutgers, the State University of New Jersey. He received his MS and PhD degrees in Computer Science from Rice University in 2004 and 2008 respectively. He was an Assistant Professor at the Department of Computer Science and Engineering at the University of Nevada, Reno from 2008 to 2012. He is the recipient of a NASA Early Career Faculty award and his research has been supported by the NSF, NASA, DHS and the DoD. His research interests include planning and coordination of robots, especially for systems with many degrees of freedom and significant dynamics, as well as applications to robotic manipulation, planetary exploration, cyber-physical systems and physically-realistic virtual agents. \end{IEEEbiography}

\begin{IEEEbiography}[{\includegraphics[width=1in]{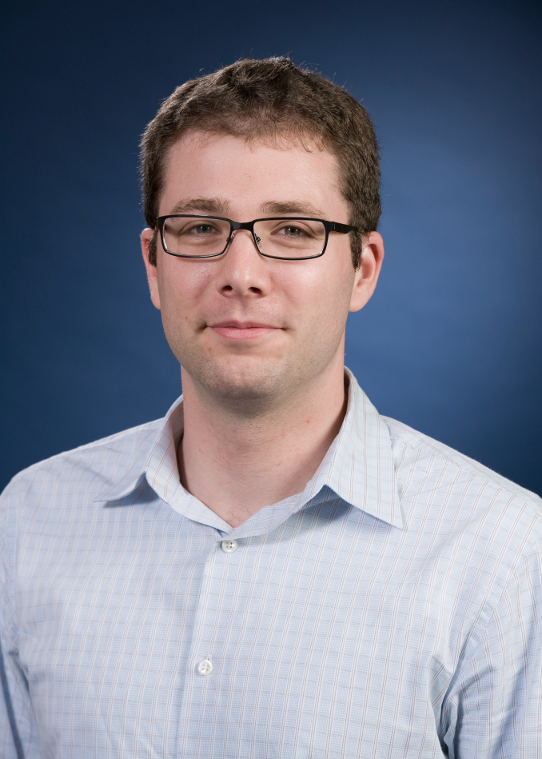}}]{Dmitry Berenson}
 received a BS in Electrical Engineering from Cornell University in 2005 and received his Ph.D. degree from the Robotics Institute at Carnegie Mellon University in 2011. He completed a post-doc at UC Berkeley in 2012 and was an Assistant Professor at WPI 2012-2016. He started as an Assistant Professor in EECS at University of Michigan in 2016. His current research focuses on motion planning, manipulation, and human-robot collaboration. He received the IEEE RAS Early Career award in 2016.
\end{IEEEbiography}

\begin{IEEEbiography}[{\includegraphics[width=1in,height=1.25in,clip,keepaspectratio]{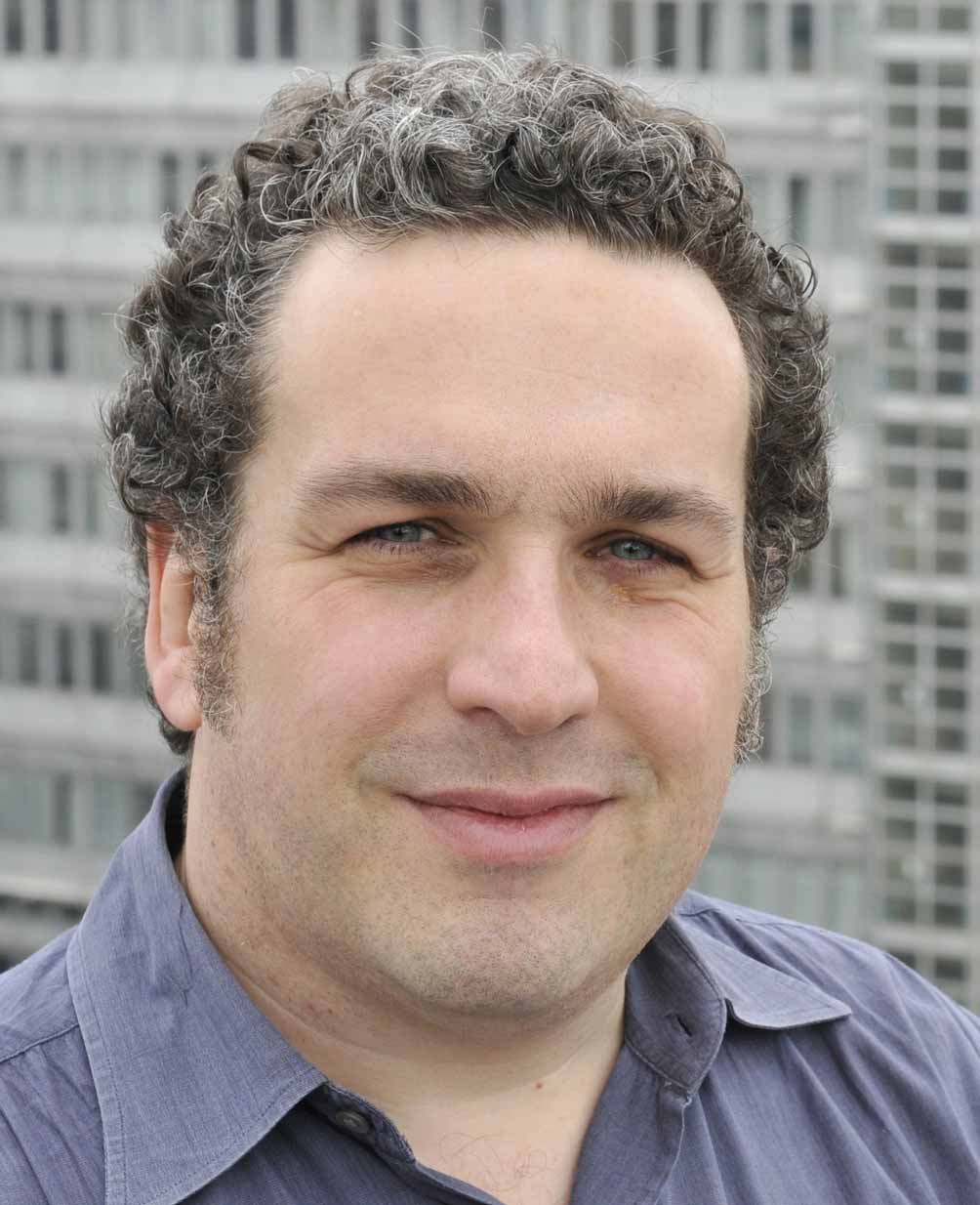}}]{Oliver Brock}
is the Alexander-von-Humboldt Professor of Robotics in the School of Electrical Engineering and Computer Science at the Technische Universit\"at Berlin in Germany. He received his Diploma in Computer Science in 1993 from the Technische Universit\"at Berlin and his Master's and Ph.D.~in Computer Science from Stanford University in 1994 and 2000, respectively. He also held post-doctoral positions at Rice University and Stanford University. Starting in 2002, he was an Assistant Professor and Associate Professor in the Department of Computer Science at the University of Massachusetts Amherst, before to moving back to the Technische Universit\"at Berlin in 2009. The research of Brock's lab, the Robotics and Biology Laboratory, focuses on autonomous mobile manipulation, interactive perception, grasping, manipulation, soft hands, interactive learning, motion generation, and the application of algorithms and concepts from robotics to computational problems in structural molecular biology. He is also the president of the Robotics: Science and Systems foundation.
\end{IEEEbiography}

\begin{IEEEbiography}[{\includegraphics[width=1in,height=1.25in,clip,keepaspectratio]{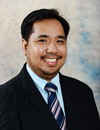}}]{Albert Causo}
is a research fellow at the Robotics Research Centre, School of Mechanical and Aerospace Engineering, Nanyang Technological University in Singapore. He received his bachelor's degree in Chemistry and Computer Engineering from Ateneo de Manila University, Philippines in 2000 and 2001, respectively. He obtained his masters and Ph.D.~in Information Science from Nara Institute of Science and Technology in 2006 and 2010, respectively. His research interests include computer vision, human-robot interaction, human motion measurement, human posture tracking and modeling, rehabilitation robotics, robot-assisted education and collaborative robots.
\end{IEEEbiography}

\begin{IEEEbiography}
[{\includegraphics[width=1in,height=1.25in,clip,keepaspectratio,trim={0.5cm 0 0.5cm 0}]{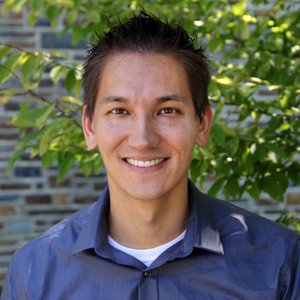}}]
{Kris Hauser} is an Associate Professor at the Pratt School of Engineering at Duke University with a joint appointment in the Electrical and Computer Engineering Department and the Mechanical Engineering and Materials Science Department. He received his PhD in Computer Science from Stanford University in 2008, bachelor's degrees in Computer Science and Mathematics from UC Berkeley in 2003, and worked as a postdoctoral fellow at UC Berkeley. He then joined the faculty at Indiana University from 2009-2014, where he started the Intelligent Motion Lab, and began his current position at Duke in 2014. He is a recipient of a Stanford Graduate Fellowship, Siebel Scholar Fellowship, Best Paper Award at IEEE Humanoids 2015, and an NSF CAREER award.  Research interests include robot motion planning and control, semiautonomous robots, and integrating perception and planning, as well as applications to intelligent vehicles, robotic manipulation, robot-assisted medicine, and legged locomotion.
\end{IEEEbiography}

\begin{IEEEbiography}
[{\includegraphics[width=1in,height=1.25in,clip,keepaspectratio]{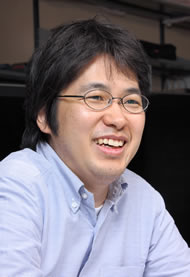}}]
{Kei Okada} is an Associate Professor in the Department of Mechano-Informatics at the University of Tokyo. He receved his BE in Computer Science from Kyoto University in 1997. He received the MS and the PhD in Information Engineering from The University of Tokyo in 1999 and 2002 respectively. 
From 2002 to 2006, he jointed the Professional Programme for Strategic Software Project in The University Tokyo.  He was appointed as a Lecturer in the Creative Informatics in 2006 and a Associate Professor in the Department of Mechano-Informatics in 2009.  His research interests include recognition-action integrated system, real-time 3D computer vision, and humanoid robots.
He is a recipient of Best Paper Award at IEEE Humanoids 2006, Best Robocup Award at IEEE/RSJ IROS 2008 and Best Conference Paper Award at IEEE ICRA 2014.
\end{IEEEbiography}

\begin{IEEEbiography}
[{\includegraphics[width=1in,height=1.25in,clip,keepaspectratio]{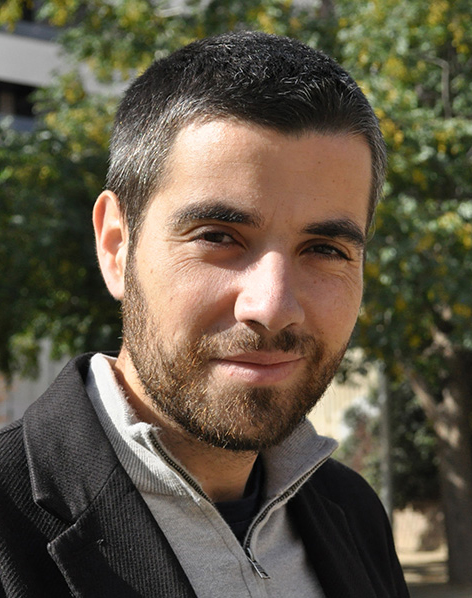}}]
{Alberto Rodriguez} is an Assistant Professor at the Mechanical Engineering Department at MIT. Alberto graduated in Mathematics ('05) and Telecommunication Engineering ('06 with honors) from the Universitat Politecnica de Catalunya (UPC) in Barcelona, and earned his PhD in Robotics (’13) from the Robotics Institute at Carnegie Mellon University. He joined  the faculty at MIT in 2014, where he started the Manipulation and Mechanisms Lab (MCube). Alberto is the recipient of the Best Student Paper Awards at conferences RSS 2011 and IEEE ICRA 2013.  His main research interests are in robotic manipulation, mechanical design, and automation.
\end{IEEEbiography}

\begin{IEEEbiography}[{\includegraphics[width=1in,height=1.25in,clip,keepaspectratio]{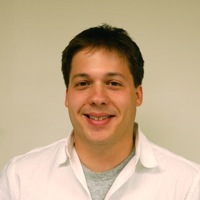}}]{Joseph Romano}
is a Research Scientist developing next-generation robotic platforms at Berkshire Grey Inc. in Waltham, Massachusetts. Prior to BG, Joe was part of the engineering team that brought the Rethink Robotics Baxter Robot to life and member of the advanced research team at Kiva Systems (Amazon Robotics) where he helped design the Amazon Picking Challenge. He received his Masters/PhD from the University of Pennsylvania$'$s GRASP laboratory focusing on algorithms for robotic tactile manipulation and virtual haptic rendering, and his BS at Johns Hopkins University working in surgical robotics. Joe$'$s interests span signal-processing, control, and planning strategies that allow robots to accomplish delicate sensor-aware manipulation tasks.
\end{IEEEbiography}

\begin{IEEEbiography}[{\includegraphics[width=1in,height=1.25in,clip,keepaspectratio]{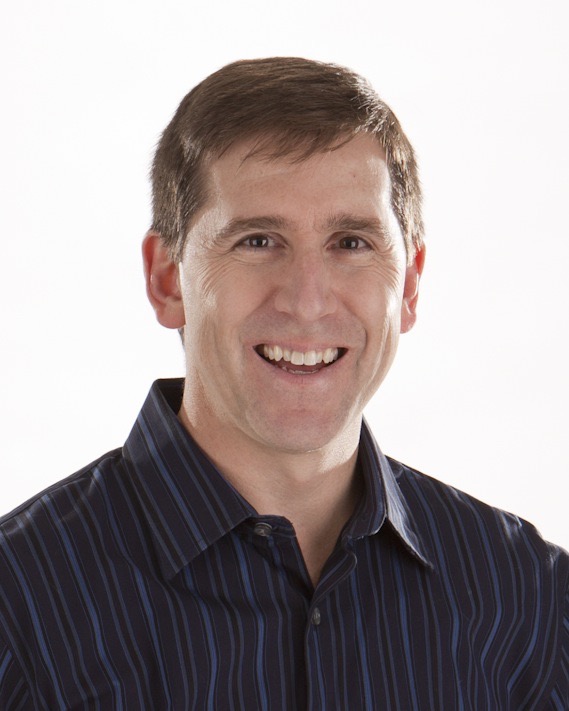}}]{Peter Wurman}
was the Chief Technology Officer and Technical Co-Founder of Kiva Systems, which was acquired by Amazon in 2012. While at Amazon, Pete conceived and organized the 2015 APC. Prior to helping found Kiva, Pete was an Associate Professor of Computer Science at North Carolina State University, member of the Operations Research Faculty, and Co-chair of the E-commerce Program. Pete received his Ph.D.~in Computer Science from the University of Michigan in 1999 where his research focused on AI and e-commerce. He received an S.B.~in Mechanical Engineering from MIT in 1987.
\end{IEEEbiography}

\end{document}